\documentclass[conference]{IEEEtran}
\IEEEoverridecommandlockouts
\usepackage{cite}
\usepackage{amsmath,amssymb,amsfonts}
\usepackage{multirow}
\usepackage{algorithmic}
\usepackage{graphicx}
\usepackage{xcolor}
\usepackage{subcaption}
\usepackage{textcomp}
\usepackage{orcidlink}
\usepackage{url}
\usepackage{hyperref}
\usepackage{titlesec}
\usepackage[nohyperlinks]{acronym}
\usepackage{changepage}
\usepackage{booktabs}
\usepackage{tabularx}
\usepackage{adjustbox}
\usepackage{wrapfig}
\usepackage{acronym}

\def\BibTeX{{\rm B\kern-.05em{\sc i\kern-.025em b}\kern-.08em
    T\kern-.1667em\lower.7ex\hbox{E}\kern-.125emX}}
\begin{document}

\title{De-SaTE: Denoising Self-attention Transformer Encoders for Li-ion Battery Health Prognostics}

\author{\IEEEauthorblockN{Gaurav Shinde \orcidlink{0009-0004-5961-9526} \IEEEauthorrefmark{1}, Rohan Mohapatra \orcidlink{0000-0003-1654-7994} \IEEEauthorrefmark{2},
Pooja Krishan \orcidlink{0000-0003-1187-6019} \IEEEauthorrefmark{2}, and Saptarshi Sengupta \orcidlink{0000-0003-1114-343X} \IEEEauthorrefmark{2}}
\IEEEauthorblockA{\IEEEauthorrefmark{1}Department of Engineering, San Jos\'e State University, San Jose, CA, USA \\
\IEEEauthorblockA{\IEEEauthorrefmark{2}Department of Computer Science, San Jos\'e State University, San Jose, CA, USA \\}
Email: \IEEEauthorrefmark{1}gauravyeshwant.shinde@sjsu.edu,
\IEEEauthorrefmark{2}rohan.mohapatra@sjsu.edu, \\
\IEEEauthorrefmark{2}pooja.krishan@sjsu.edu,
\IEEEauthorrefmark{2}saptarshi.sengupta@sjsu.edu}}


\maketitle
\begin{abstract}
The usage of \ac{Li-ion} batteries has gained widespread popularity across various industries, from powering portable electronic devices to propelling electric vehicles and supporting energy storage systems. A central challenge in \ac{Li-ion} battery reliability lies in accurately predicting their \ac{RUL}, which is a critical measure for proactive maintenance and predictive analytics.
This study presents a novel approach that harnesses the power of multiple denoising modules, each trained to address specific types of noise commonly encountered in battery data. Specifically, a denoising auto-encoder and a wavelet denoiser are used to generate encoded/decomposed representations, which are subsequently processed through dedicated self-attention transformer encoders. After extensive experimentation on NASA and CALCE data, a broad spectrum of health indicator values are estimated under a set of diverse noise patterns. The reported error metrics on these data are on par with or better than the state-of-the-art reported in recent literature. 
\end{abstract}


\section{Introduction}
Lithium-ion batteries (\ac{Li-ion}) are the leading energy storage solution, prized for their exceptional energy density, rapid power response, recyclability, and portability. Their unparalleled combination of energy and power density has made them the preferred choice for applications ranging from hybrid and electric vehicles to portable electronics. But, \ac{Li-ion} battery capacity can deteriorate with time, influenced by factors such as temperature, state of charge, cycling rate, and operating conditions, leading to reduced performance and potential failure. The battery capacity is a key health indicator, crucial for accurately forecasting the Remaining Useful Life (\ac{RUL}) before reaching \ac{EOL} when performance significantly degrades or rated capacity can no longer be sustained.



\ac{PHM} encompasses data acquisition, diagnostics, and the core component of prognostics, which predicts a system's \ac{RUL}. In the context of \ac{Li-ion} batteries, \ac{PHM} provides critical insights into monitoring their health, guiding maintenance decisions, and reducing the risk of unexpected failures, particularly in safety-critical applications like electric vehicles and aerospace systems.

In the past, \ac{RUL} prediction primarily relied on conventional machine learning models such as \ac{CNN}\cite{CNN}, \ac{RNN}\cite{RNN}, and \ac{LSTM}\cite{LSTM} networks. These models, while valuable, often faced challenges in capturing long-term dependencies in sequential data and required manual feature engineering. Consequently, they are less flexible in handling diverse datasets and complex real-world scenarios.

Recent \ac{RUL} prediction trends focus on attention-based mechanisms, known for autonomously capturing intricate temporal and spatial data dependencies, reducing the requirement for extensive feature engineering. This shift away from traditional approaches aims to boost prediction accuracy and adaptability in complex systems\cite{shift}. Chen et. al \cite{1} explore the utilization of a Denoising Auto-Encoder to enhance data representation from battery inputs, which inherently exhibit noise due to various factors. However, it's essential to acknowledge that measurement noise in practical scenarios may not adhere to only a Gaussian distribution as assumed here. To address this concern and ensure robust noise handling, this research proposes a novel approach. Diverse set of noise types are implemented in the denoising framework, where each noise type is associated with a dedicated auto-encoder and its corresponding transformer encoder. The decoded representations from these auto-encoders undergo transformation through the respective transformer encoders. Then, a minimization layer which identifies the noise type that yields the minimum error value is introduced. This adaptive noise modeling approach bolsters the auto-encoder's capacity to capture the spectrum of non-Gaussian noise characteristics commonly encountered in battery data. Due to the denoising process, this method culminates in generating higher-quality data representations for subsequent analysis, where the prediction is based on the noise type associated with the minimum error.


\hspace{0.2 cm}\textbf{Contributions:}
The major contributions of this paper are:
\begin{enumerate}
\item \textbf{Multi-Faceted Noise Mitigation:}
   This work introduces a comprehensive noise mitigation strategy by employing dedicated denoising auto-encoders and wavelet decomposers for various noise types present in battery operational data. Each denoising module is tailored to handle specific noise characteristics, enhancing the model's adaptability and relevance in real-world scenarios.

   \item \textbf{Robustness to different magnitudes of noise:}
  The proposed denoisers are tuned to handle various noise levels and noise distributions, thus making the architecture more robust to fluctuations in the input.

\item \textbf{Better data representation leading to enhanced accuracy:}
    This architecture effectively denoises the input data and improves the quality of data representations leading to better predictions.
   
\item \textbf{A modular architecture for all complex processing:}
   The proposed architecture processes the input by passing it through the denoising modules to encode various types of noisy data, and the self-attention encoder network subsequently learns the degradation physics and predicts the remaining useful life.

\end{enumerate}

\section{Self-Attention with Variable Denoising}
\begin{figure}[t!]
  \centering
      \includegraphics[width=1\columnwidth]{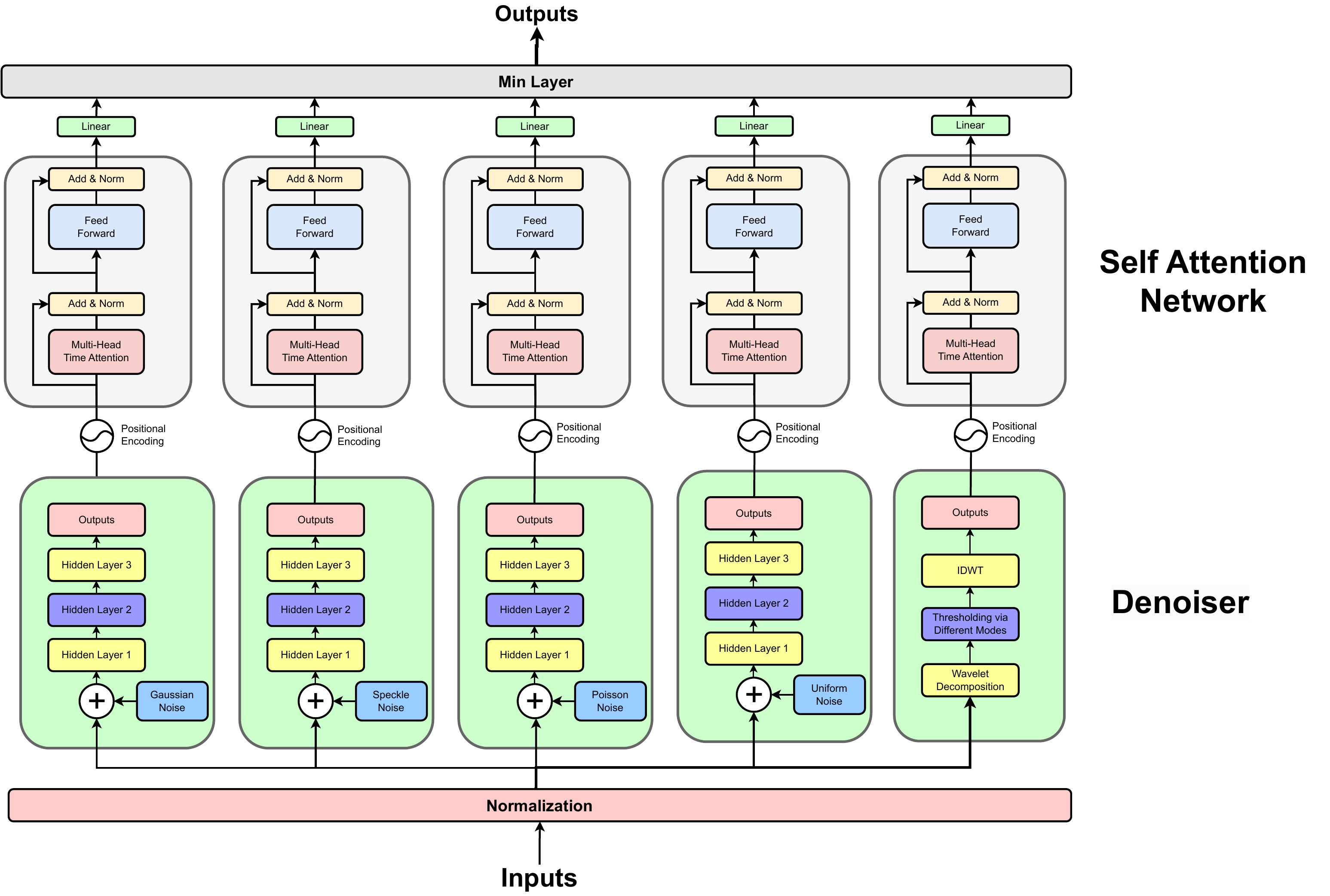}
      \caption{The proposed \textbf{De-SaTE}: Denoising Self-attention Transformer Encoder architecture}
    \label{system_architecture}
      \vspace*{-0.15in}
\end{figure}

In the input data $x = \{x_1, x_2, \ldots, x_n\}$, where $ x \in (0, 1]$. A normalized input is produced, $x' = \frac{x}{C_0}$ where $C_0$ denotes rated capacity. Subsequently, multiple denoising schemes, each trained to mitigate a specific noise type are leveraged. The encoded representations from these denoising modules are then subjected to individual self-attention layers. The data passes through the self-attention layers, each one intricately connected to its respective noise reducer, which then yields individual metric values such as Relative Error (RE), Mean Absolute Error (MAE), and Root Mean Square Error (RMSE). A minimization strategy is employed, which selects the minimum value among these metrics, thereby obtaining the most optimal error estimate.
This architecture ensures robust performance and is highly effective in estimating failures, even when confronted with the presence of diverse noise types inherent in battery data. The system architecture of the entire process is shown in Fig.~\ref{system_architecture}. 

\subsection{Positional Encoding}
\begin{equation}
PE_{(pos, 2k)} = \sin\left(\frac{pos}{10000^{2k/d_{\text{k}}}}\right)
\end{equation}

\begin{equation}
PE_{(pos, 2k+1)} = \cos\left(\frac{pos}{10000^{2k/d_{\text{k}}}}\right)
\end{equation}

\subsection{Self Attention}
\begin{figure}[htbp]
\centering
\begin{subfigure}[b]{\columnwidth}
        \centering
        \includegraphics[width=0.8\columnwidth]{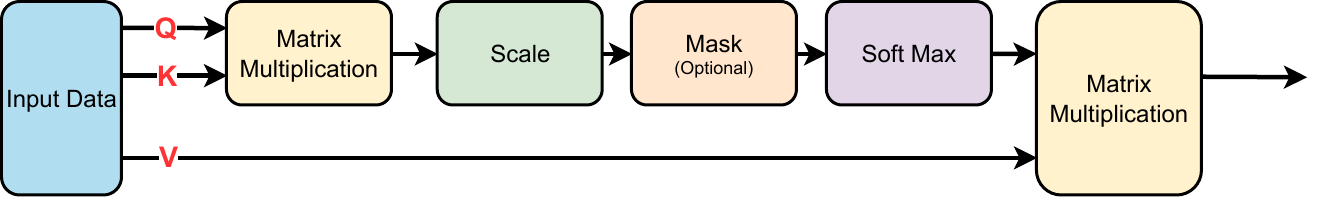}
   \caption{Scaled Dot-Product Self-Attention mechanism}
   \label{fig:scaleddot}
\end{subfigure}
\\~\\
\begin{subfigure}[b]{\columnwidth}
\centering
   \includegraphics[width=0.8\columnwidth]{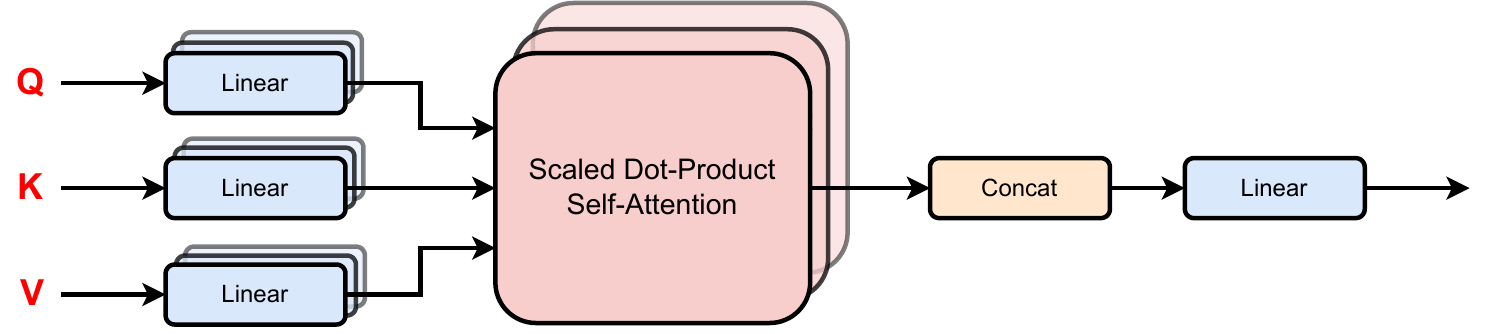}
   \caption{Multi-Head Self-Attention mechanism}
   \label{fig:multihead}
\end{subfigure}
\caption{Multi-Head Self-Attention architecture}
\label{fig:Ng2}
\vspace*{-0.15in}
\end{figure}

The encoder's self-attention mechanism \cite{vaswani_trans} computes attention scores for each position in the input sequence, allowing the model to weigh the importance of different elements in the sequence when encoding a particular position. This is typically computed using a weighted sum of queries, keys and values as shown Fig. \ref{fig:scaleddot}.

\begin{equation}
    \text{Attention}(Q, K, V) = \text{softmax} (\frac{Q K'}{\sqrt{d_{k}}}) V
\end{equation}

Here, $Q$  represents the query matrix for the input sequence, $K$ represents the key matrix for the input sequence, $V$ represents the value matrix for the input sequence and $d_k$ is a scaling factor to stabilize the gradients.

\subsection{Multi-head Attention}
In Fig. \ref{fig:multihead},  the multi-head attention process is outlined. The self-attention mechanism is often used in multiple heads \cite{vaswani_trans} to capture different types of dependencies:
\begin{equation}
\text{MultiHead}(Q,K,V) = \text{Concat}(\text{head}_1, \text{head}_2, ..., \text{head}_h)\cdot W^O
\end{equation}
Here, $\text{head}_i$ represents the output of the $i$-th attention head and $W^O$ is a learnable weight matrix.

\subsection{Feed forward network}
After computing attentions, the output passes through a Feed-Forward Neural network:
\begin{equation}
    \text{FFN}(x) = \text{ReLU}(W_1 \cdot x + b_1)\cdot W_2 + b_2
\end{equation}
Here, $W_1$, $b_1$, $W_2$, $b_2$ are learnable weights and biases.

\subsection{Learning}
The proposed architecture can be divided into two tasks: \textit{denoising} and \textit{metric evaluation}. The learning procedure optimizes both tasks simultaneously within a unified framework. Mean Square Error (MSE) is used to evaluate loss, and the objective function $L$\cite{1} is defined as follows:.
\begin{equation}
L = \sum_{t=T+1}^{n} (x_t - \hat{x_t})^2 + \delta \sum_{i=1}^{n} \ell(x_{corr} - \hat{x_i}) + \alpha \psi(L_{rate})
\label{eqn:loss}
\end{equation}

where, $n$ is the number of samples, $\delta$ controls relative contribution of each task, $\ell(\cdot)$ is the loss function, $\alpha$ is a regularization parameter, $\psi(\cdot)$ denotes regularization and $L_{rate}$ denotes learning parameters.

The denoising effect along with penalized loss acts like a regularizer. Regularization techniques \cite{Kress1989} add a penalty term to the loss function, boosting model performance to tend towards smaller weights or simpler representations.

\section{Experimental Setup}
\subsection{Dataset Description}
\begin{figure}[htbp]
\hspace{-0.2in}
\centering
\begin{subfigure}[t]{0.45\columnwidth}
        \centering
        \includegraphics[width=1.1\columnwidth]{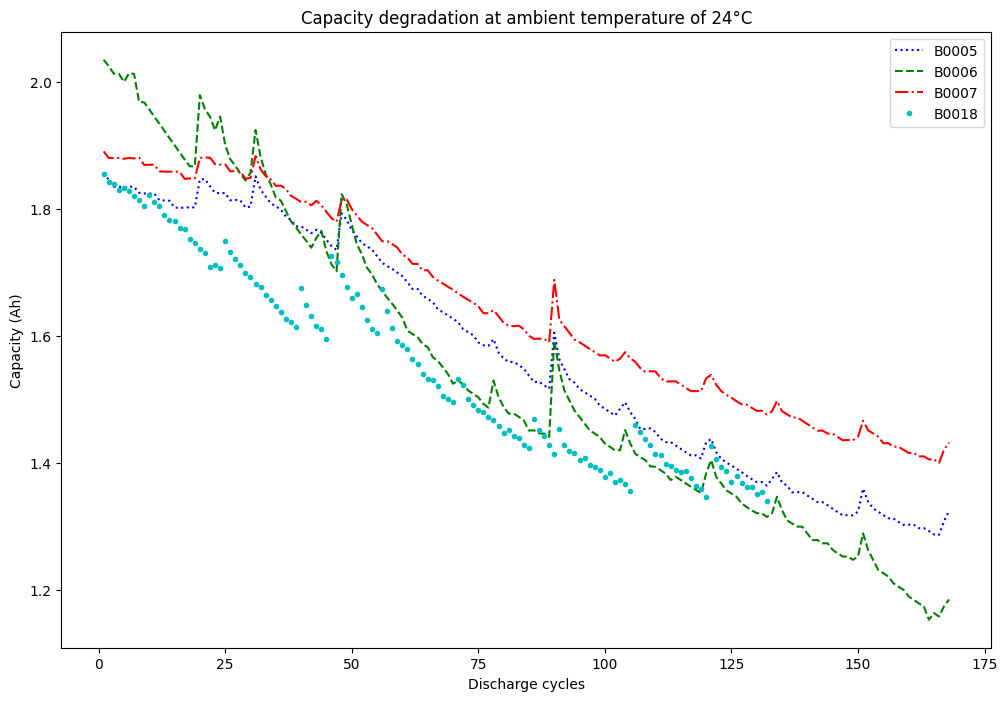}
   \caption{Degradation trend on the NASA dataset}
   \label{NASA}
\end{subfigure}
~
\begin{subfigure}[t]{0.45\columnwidth}
\centering
   \includegraphics[width=1.1\columnwidth]{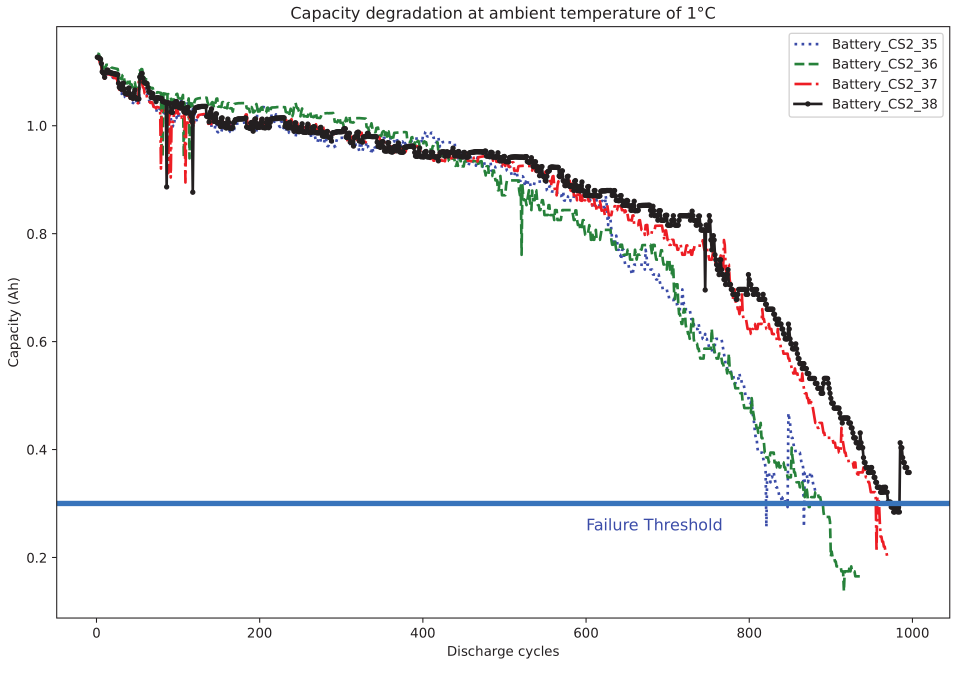}
   \caption{Degradation trend on the CALCE dataset}
   \label{CALCE}
\end{subfigure}
\caption{Capacity vs. degradation cycles}
\label{NASAaCALCE}
\vspace{-0.15in}
\end{figure}

Two datasets from \ac{NASA} and \ac{CALCE} were used to conduct the experiments. The \ac{NASA} dataset, acquired from the \ac{NASA} Ames Research Center, comprises of records from four different Li-ion batteries (B0005, B0006, B0007 and B0018), each subjected to three distinct operations: charging, discharging, and impedance measurements \cite{83, 84}. The \ac{CALCE} dataset (CS2\_35, CS2\_36, CS2\_37 and CS2\_38) is sourced from the Center for Advanced Life Cycle Engineering (CALCE) at the University of Maryland \cite{85}. Figures \ref{NASA} and \ref{CALCE} illustrate the capacity degradation trends observed across various batteries in these datasets.


\subsection{Noise distributions}
\subsubsection{Gaussian Noise}
Gaussian noise, characterized by mean ($\mu$) and standard deviation ($\sigma$) has the \ac{PDF}:
\begin{equation}
    f(x;\mu,\sigma) = \frac{1}{\sigma\sqrt{2\pi}}e^{-\frac{(x-\mu)^2}{2\sigma^2}}
\end{equation}
\subsubsection{Speckle Noise}

Speckle noise is often multiplicative, where input values are multiplied by random values. Its \ac{PDF} is:

\begin{equation}
    f(x;\gamma) = \frac{1}{\gamma^2}e^{-\frac{x}{\gamma}}
\end{equation}
where $\gamma$ is a parameter controlling the noise intensity.

\subsubsection{Poisson Noise}
Poisson noise is characterized by its mean ($\lambda$). Poisson noise manifests when events occur at a consistent average rate but with randomness in the exact timing or occurrence of these events. This noise, explained as instrumentation noise in battery health prognosis has the \ac{PDF}:
\begin{equation}
    f(x;\lambda) = \frac{e^{-\lambda}\lambda^x}{x!}
\end{equation}
\subsubsection{Uniform Noise}
Uniform noise characterized by a minimum value ($a$) and a maximum value ($b$), has the \ac{PDF}:
\begin{equation}
f(x;a,b) = \begin{cases}
\frac{1}{b-a} & \text{if } a \leq x \leq b \\
0 & \text{otherwise}
\end{cases}
\end{equation}

\subsection{Denoising Autoencoder}

A denoising autoencoder is effective in removing noise from data and learning robust representations. The network is trained to reconstruct clean data from noisy input. It consists of an encoder that maps the input data to a latent representation and a decoder that reconstructs the data from this representation. During training, the encoder learns to capture essential features while the decoder learns to remove noise. The loss function defined in Eqn. \ref{eqn:loss} typically measures reconstruction error, encouraging the network to minimize differences between clean inputs and reconstructed outputs.


Let $x_{0t} = x_{0(t+1)}, x_{0(t+2)}, \ldots, x_{0(t+m)} \in x_{0}$ denote the slice of input with $m$ samples of a sequence. A noise is added to the normalized input to obtain the corrupted vector, $x_{corr}$.

DAE serves two purposes: denoising the raw input and learning a nonlinear representation \cite{1}:
\begin{equation}
    z = a\left(W\cdot x_{corr} + b\right)
\end{equation}
where $W$, $b$, $a(\cdot)$, and $z$ denote weight, bias, activation function, and the output of the DAE encoder.

Then, to reconstruct the input vector, the latent representation is mapped back to the input space, defined as follows:
\begin{equation}
    \hat{x}_t = f_0\left(W_0\cdot z + b_0\right)
\end{equation}
where $W_0$, $b_0$, $z$, and $f_0(\cdot)$ denote weight, bias, output, and map function of the output layer of the DAE encoder.

In this network, identity and ReLU functions are used as the decoding and encoding activation, respectively. Finally, the objective function $L_d$ is defined as follows:
\begin{equation}
    L_d = \frac{1}{n} \sum_{t=1}^{n} \left(x_{corr} - \hat{x}_t\right)^2 + \alpha \left(\|W\|_F^2 + \|W_0\|_F^2\right)
\end{equation}
where $\|\cdot\|_F$ is the Fibonacci-norm, $\alpha$ is the regularization parameter, and $n$ is the number of samples.

\subsection{Wavelet Transformation and Denoiser}
 Wavelet denoising is a signal processing technique for removing noise from a voltage signal coming from a battery or another source. Wavelet denoising \cite{CHO200577} typically involves thresholding coefficients obtained from wavelet transforms. A typical transformation involves passing the signal through \ac{DWT}, thresholding the wavelet coefficients. The denoised signal is reconstructed using the inverse DWT. The process is outlined below.

 \subsubsection{\ac{DWT}}
        The \ac{DWT} decomposes a signal or image into wavelet coefficients at different scales and positions:
        
        \begin{equation}
            \mathbf{\theta} = \text{\ac{DWT}}(I)
        \end{equation}
        
        $I$ is the original signal, $\mathbf{\theta}$ contains wavelet coefficients.
        
        \subsubsection{Thresholding}
        Thresholding is applied to the wavelet coefficients to remove or reduce noise. A common method is \textit{Soft} thresholding:
        \begin{equation}
            \hat{\theta}_{i,j} = \text{sign}(\theta_{i,j}) \cdot \max\left(|\theta_{i,j}| - \epsilon, 0\right)
        \end{equation}
        
        Another approach to thresholding is \textit{Hard} thresholding. Hard thresholding sets coefficients below a certain threshold to zero and retains those above the threshold. It is defined as follows:
        \begin{equation}
            \hat{\theta}_{i,j} = \begin{cases}
                \theta_{i,j}, & \text{if } |\theta_{i,j}| \geq \epsilon \\
                0, & \text{if } |\theta_{i,j}| < \epsilon
            \end{cases}
        \end{equation}
        
        An additional thresholding method, \textit{Garrote}, is a variation penalizing large coefficients than smaller ones and is given as:
        \begin{equation}
            \hat{\theta}_{i,j} = \frac{\text{sign}(\theta_{i,j}) \cdot \max\left(|\theta_{i,j}| - \epsilon, 0\right)}{1 + \frac{\epsilon}{|\theta_{i,j}|}}
        \end{equation}
        
        where, $\hat{\theta}_{i,j}$ is the denoised coefficient, $\theta_{i,j}$ is the original coefficient, and $\epsilon$ is the threshold value.
        
        \subsubsection{\ac{IDWT}}
        
        \ac{IDWT} reconstructs signals using this transformation:
        
        \begin{equation}
            \hat{I} = \text{\ac{IDWT}}(\hat{\mathbf{\theta}})
        \end{equation}
        
        $\hat{I}$ is the denoised signal, $\hat{\mathbf{\theta}}$ contains the denoised wavelet coefficients.


\subsection{Training and Evaluation}
Four types of noise were introduced: Gaussian, Speckle, Poisson, and Uniform with varying noise levels (small, medium, and relatively high). The optimal hyperparameters are evaluated by a grid search.
\vspace{0.1 cm}
\begin{itemize}
    \item \textit{Learning Rate:} 1e-3 and 1e-2
    \item \textit{Number of Layers:} 1 and 2
    \item \textit{Hidden Dimension:} 16 and 32
    \item \textit{Noise Levels:} 0.001, 0.01, and 0.05 
    \item \textit{Epochs:} 2000
\end{itemize}

\begin{figure*}[t!]
    \centering
    \begin{subfigure}[t]{0.5\textwidth}
        \centering
        \includegraphics[width=1\textwidth]{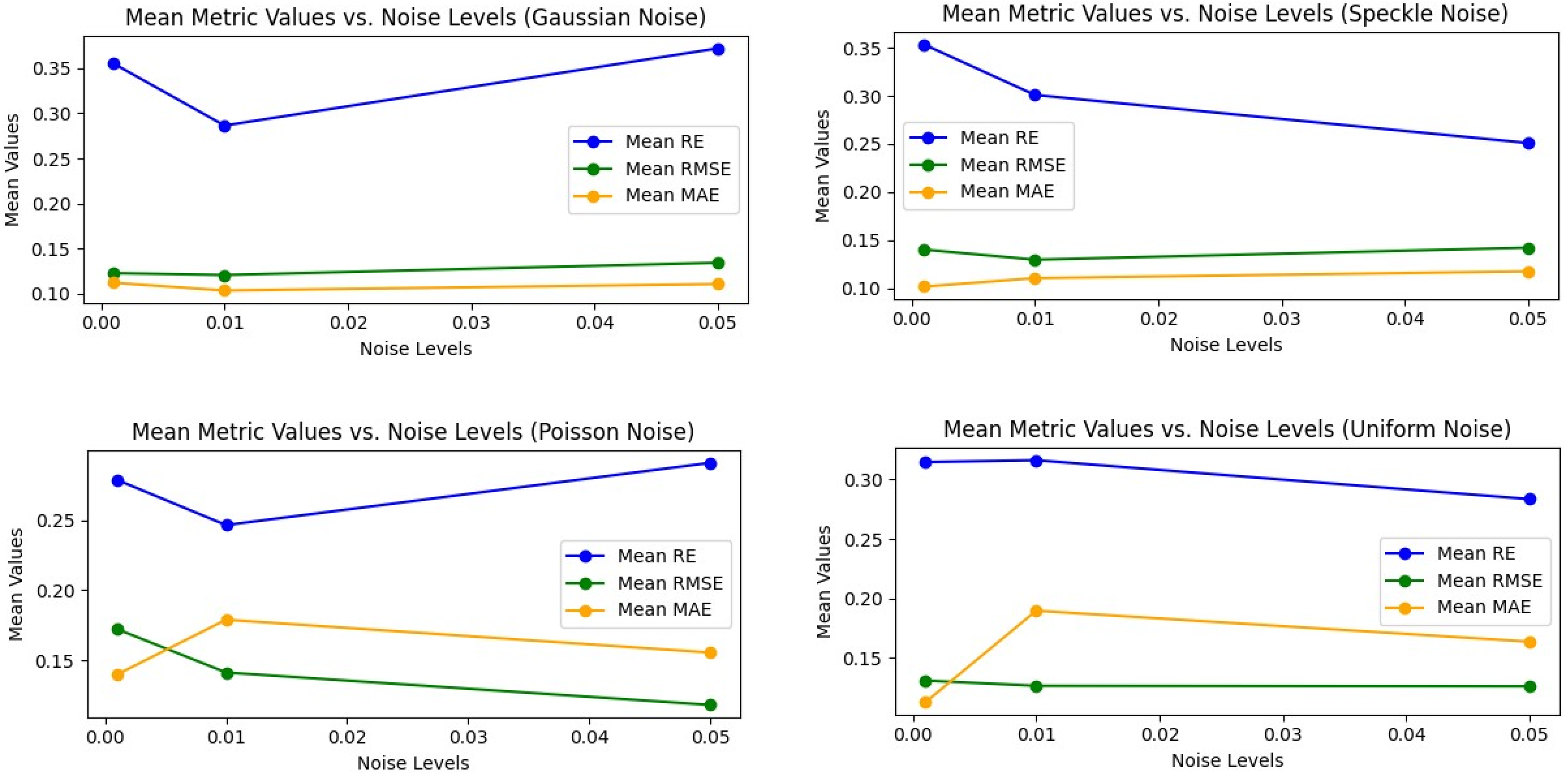}
        \caption{Mean \ac{RE}, \ac{RMSE}, \ac{MAE} values on the NASA data under different types of noise with varying levels}
        \label{results-nasa}
    \end{subfigure}%
    ~
    \begin{subfigure}[t]{0.5\textwidth}
        \centering
        \includegraphics[width=0.9\textwidth]{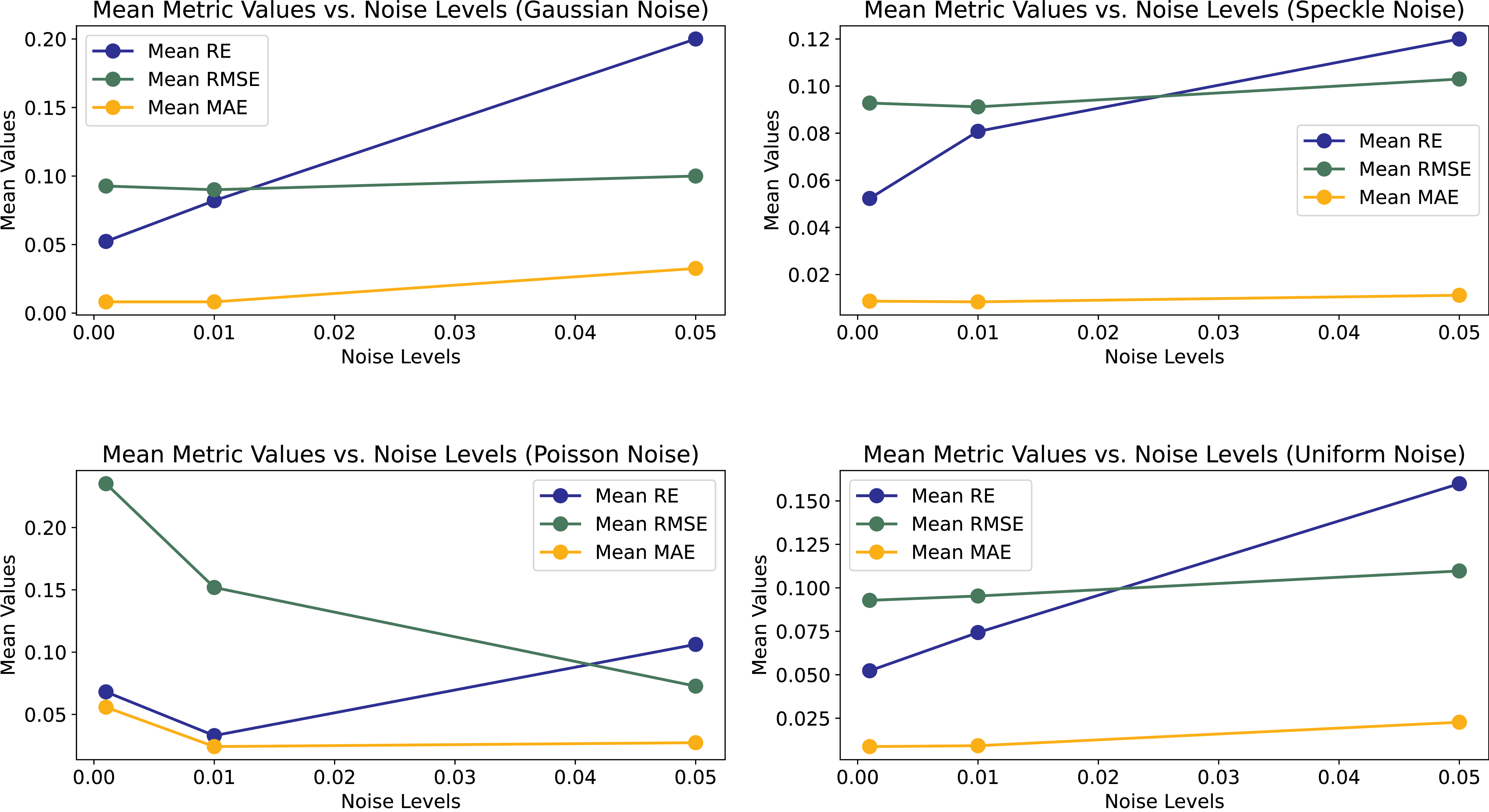}
        \caption{Mean \ac{RE}, \ac{RMSE}, \ac{MAE} values on the CALCE data under different types of noise with varying levels}
         \label{results-calce}
    \end{subfigure}
    \caption{Comparison of metrics under diverse types and levels of noise}
\end{figure*}

The experiments were run on a system using Python 3.10, Tensorflow 2.0 and Keras with Nvidia A100 and Nvidia T4 GPUs. 

The models are evaluated using three key metrics outlined in Appendix \ref{SecondAppendix}. This systematic exploration is aimed to identify the best hyperparameters for accurate predictions amidst diverse noise distributions, ensuring robustness and scalability.

\section{Results}

Figures \ref{results-nasa} and \ref{results-calce} demonstrate the \ac{RE}, \ac{MAE}, and \ac{RMSE} values under various noise distributions for the \ac{NASA} and \ac{CALCE} datasets. A comparative analysis as shown in Fig. \ref{fig:Ng2} of how each noise affects the key metrics used in this paper is performed. The results are tabulated in Tables \ref{tab1} and \ref{tab2}.



\begin{figure}[htbp]
\hspace{-0.1in}
\centering
\begin{subfigure}[t]{0.48\columnwidth}
        \centering
        \includegraphics[width=1.05\columnwidth]{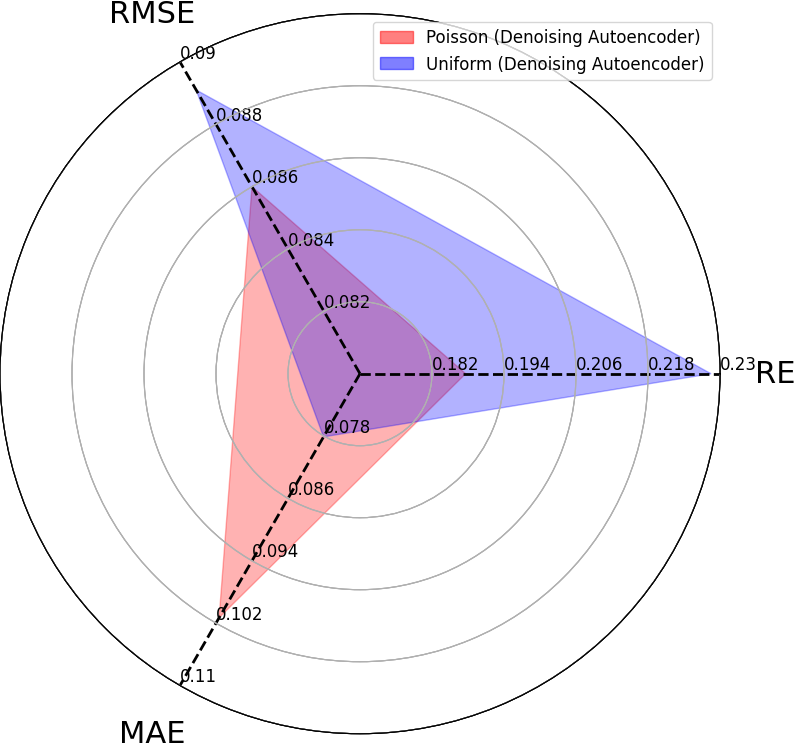}
   \caption{Poisson noise with lower RE and RMSE but higher MAE than Uniform noise on NASA Dataset (evaluated with Denoising Autoencoder)}
   \label{fig:poissonNasa}
\end{subfigure}
~
\begin{subfigure}[t]{0.48\columnwidth}
\centering
   \includegraphics[width=1.05\columnwidth]{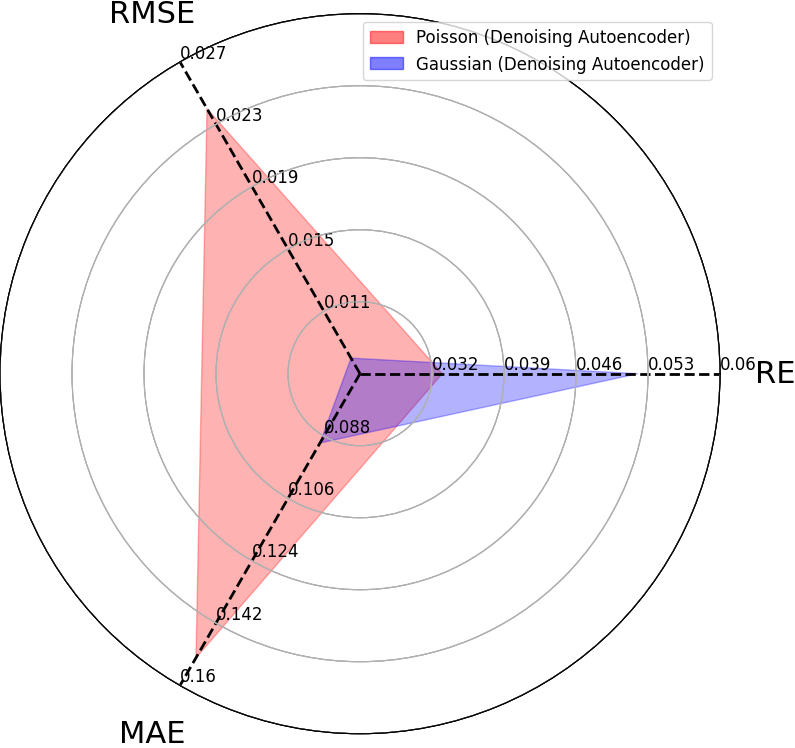}
   \caption{Poisson noise with lower RE but higher RMSE and MAE than Gaussian noise on CALCE Dataset (evaluated with Denoising Autoencoder)}
   \label{fig:poissonCalce}
\end{subfigure}
\vspace{0.1 cm}
\caption{Effects of different noise on metrics (RE, RMSE, MAE)}
\label{fig:Ng2}
\end{figure}

\begin{table}[htpb!]
\centering
\caption{Results for the NASA dataset}
\label{tab1}
\begin{tabular}{ll|llllll}
\multicolumn{2}{l|}{Noise and metrics}& LR & NoL & HD & $\alpha$ & NL & Result \\ \hline
\multicolumn{1}{l|}{\multirow{3}{*}{Gaussian}} & RE & 0.01 & 1 & 16 & 1e-05 & 0.05 & 0.1674 \\  
\multicolumn{1}{l|}{} & MAE & 0.01 & 1 & 16 & 1e-05 & 0.05 & 0.0806 \\ 
\multicolumn{1}{l|}{} & RMSE & 0.01 & 2 & 16 & 1e-05 & 0.01 & 0.0957 \\ \hline
\multicolumn{1}{l|}{\multirow{3}{*}{Speckle}} & RE & 0.01 & 1 & 16 & 0.0001 & 0.05 & 0.1869 \\  
\multicolumn{1}{l|}{} & MAE & 0.01 & 1 & 16 & 1e-05 & 0.01 & 0.0807 \\ 
\multicolumn{1}{l|}{} & RMSE & 0.01 & 1 & 16 & 1e-05 & 0.01 & 0.0935 \\ \hline
\multicolumn{1}{l|}{\multirow{3}{*}{Poisson}} & RE & 0.01 & 2 & 16 & 0.0001 & 0.01 & 0.1876 \\  
\multicolumn{1}{l|}{} & MAE & 0.01 & 2 & 16 & 1e-5 & 0.05 &  0.0860\\ 
\multicolumn{1}{l|}{} & RMSE & 0.01 & 2 & 32 & 1e-5 & 0.001 & 0.1013\\ \hline
\multicolumn{1}{l|}{\multirow{3}{*}{Uniform}} & RE & 0.001 & 2 & 32 & 1e-5 & 0.05 & 0.2285 \\  
\multicolumn{1}{l|}{} & MAE & 0.001 & 1 & 32 & 0.0001 & 0.001 & 0.0891 \\ 
\multicolumn{1}{l|}{} & RMSE & 0.001 & 1 & 32 & 0.0001 & 0.001 & \textbf{0.0781}\\ \hline
\end{tabular}
\vspace*{-0.15in}
\end{table}

\begin{table}[htpb!]
\vspace{0.25in}
\centering
\caption{Results for the CALCE dataset}
\label{tab2}
\begin{tabular}{ll|llllll}
\multicolumn{2}{l|}{Noise and metrics}& LR & NoL & HD & $\alpha$ & NL & Result \\ \hline
\multicolumn{1}{l|}{\multirow{3}{*}{Gaussian}} & RE & 0.001 & 1 & 32 & 0.01 & 0.001 & 0.052 \\  
\multicolumn{1}{l|}{} & MAE & 0.001 & 1 & 32 & 0.01 & 0.01 & \textbf{0.008} \\ 
\multicolumn{1}{l|}{} & RMSE & 0.001 & 1 & 32 & 0.01 & 0.01 & 0.09 \\ \hline
\multicolumn{1}{l|}{\multirow{3}{*}{Speckle}} & RE & 0.001 & 1 & 32 & 0.01 & 0.001 & 0.052 \\  
\multicolumn{1}{l|}{} & MAE & 0.001 & 1 & 32 & 0.01  & 0.01 & \textbf{0.008}\\ 
\multicolumn{1}{l|}{} & RMSE & 0.001 & 1 & 32 & 0.01 & 0.01 & 0.091 \\ \hline
\multicolumn{1}{l|}{\multirow{3}{*}{Poisson}} & RE & 0.001 & 1 & 32 & 0.01 & 0.01 & 0.033 \\  
\multicolumn{1}{l|}{} & MAE & 0.001 & 1 & 32 & 0.01 & 0.01 &  0.024\\ 
\multicolumn{1}{l|}{} & RMSE & 0.001 & 1 & 32 & 0.01 & 0.01 & 0.152\\ \hline
\multicolumn{1}{l|}{\multirow{3}{*}{Uniform}} & RE & 0.001 & 1 & 32 & 0.01 & 0.001 & 0.052 \\  
\multicolumn{1}{l|}{} & MAE & 0.001 & 1 & 32 & 0.01 & 0.001 & 0.009 \\ 
\multicolumn{1}{l|}{} & RMSE & 0.001 & 1 & 32 & 0.01 & 0.001 & 0.093\\ \hline
\end{tabular}
\end{table}
The network performed optimally on the \ac{NASA} dataset with a \ac{LR} of 0.01, \ac{NoL} of 1, 16 \ac{HD}, and an $\alpha$ of 1e-05. In contrast, the optimal parameters for the \ac{CALCE} dataset involve a \ac{LR} of 0.001, a \ac{NoL} of 1, 32 \ac{HD}, and an $\alpha$ of 0.01. These configurations were fine-tuned for the auto-encoder and transformer encoder layers via a grid search.



\ac{RE} is highly related to the \ac{RUL} of a battery, and serves as the primary evaluation metric. After thorough experimentation, it is observed that \ac{RE} achieved superior results when paired with a denoising autoencoder followed by a transformer encoder. The three wavelet denoising modes - Soft, Hard, and Garrote - each with three distinct thresholds (0.001, 0.01, and 0.05) are explored to comprehensively assess their impact on the overall performance. Results are tabulated in Table \ref{tab:wavelet}. 

\begin{table}[htbp]
\centering
\caption{Mean RE distribution for different wavelet modes and thresholds}
\label{tab:wavelet}
\scalebox{1.2}{\begin{tabular}{l|l|ccc}
Dataset & \multicolumn{1}{c|}{Wavelet Denoising Mode} & \multicolumn{3}{c}{Threshold} \\ 
\hline
\textbf{} & \multicolumn{1}{c|}{\textbf{}} & \multicolumn{1}{c}{0.001} & \multicolumn{1}{c}{0.01} & 0.05 \\ \hline
\multicolumn{1}{c|}{\multirow{3}{*}{\textbf{NASA}}} & Soft & \multicolumn{1}{c}{0.29} & \multicolumn{1}{c}{0.31} & 0.52 \\ 
\multicolumn{1}{c|}{} & Hard & \multicolumn{1}{c}{0.213} & \multicolumn{1}{c}{0.12} & 0.17 \\ 
\multicolumn{1}{c|}{} & Garotte & \multicolumn{1}{c}{0.24} & \multicolumn{1}{c}{0.27} & 0.22 \\ \hline
\multirow{3}{*}{\textbf{CALCE}} & Soft & \multicolumn{1}{c}{0.76} & \multicolumn{1}{c}{0.78} & \multicolumn{1}{c}{0.81} \\ 
 & Hard & \multicolumn{1}{c}{0.56} & \multicolumn{1}{c}{0.61} & \multicolumn{1}{c}{0.62} \\ 
 & Garotte & \multicolumn{1}{c}{0.65} & \multicolumn{1}{c}{0.67} & \multicolumn{1}{c}{0.71} \\ \hline
\end{tabular}}
\end{table}


\begin{table}[htbp]
\centering
\caption{Model Evaluation on \ac{Li-ion} Battery Datasets}
\label{Model evaluation}
\scalebox{1.2}{\begin{tabular}{l|l|ccc}
Dataset & \multicolumn{1}{c|}{Model} & \multicolumn{3}{c}{Metrics} \\ 
\hline
\textbf{} & \multicolumn{1}{c|}{\textbf{}} & \multicolumn{1}{c}{\textbf{RE}} & \multicolumn{1}{c}{\textbf{MAE}} & \textbf{RMSE} \\ \hline
\multicolumn{1}{c|}{\multirow{6}{*}{\textbf{NASA}}} & MLP \cite{MLP} & \multicolumn{1}{c}{0.3851} & \multicolumn{1}{c}{0.1379} & 0.1541 \\ 
\multicolumn{1}{c|}{} & RNN \cite{RNN} & \multicolumn{1}{c}{0.2851} & \multicolumn{1}{c}{0.0749} & 0.0848 \\ 
\multicolumn{1}{c|}{} & LSTM \cite{LSTM} & \multicolumn{1}{c}{0.2648} & \multicolumn{1}{c}{0.0829} & 0.0905 \\ 
\multicolumn{1}{c|}{} & GRU \cite{GRU} & \multicolumn{1}{c}{0.3044} & \multicolumn{1}{c}{0.0806} & 0.0921 \\ 
\multicolumn{1}{c|}{} & Dual-LSTM \cite{DUAL-LSTM} & \multicolumn{1}{c}{0.2557} & \multicolumn{1}{c}{0.0815} & 0.0879 \\
\multicolumn{1}{c|}{} & DeTransformer \cite{1} & \multicolumn{1}{c}{0.2252} & \multicolumn{1}{c}{\textbf{0.0713}} & 0.0802 \\
\multicolumn{1}{c|}{} &\textbf{De-SaTE} & \multicolumn{1}{c}{\textbf{0.1674}} & \multicolumn{1}{c}{0.0806} & \textbf{0.0781} \\
\hline
\specialrule{0.1em}{0.05em}{0.05em}
\multirow{6}{*}{\textbf{CALCE}} & MLP \cite{MLP} & \multicolumn{1}{c}{0.4018} & \multicolumn{1}{c}{0.1557} & \multicolumn{1}{c}{0.2038} \\ 
\multicolumn{1}{c|}{} & RNN \cite{RNN} & \multicolumn{1}{c}{0.1614} & \multicolumn{1}{c}{0.0938} & 0.1099\\ 
\multicolumn{1}{c|}{} & LSTM \cite{LSTM} & \multicolumn{1}{c}{0.0902} & \multicolumn{1}{c}{0.0582} & 0.0736 \\ 
\multicolumn{1}{c|}{} & GRU \cite{GRU} & \multicolumn{1}{c}{0.1319} & \multicolumn{1}{c}{0.0671} & 0.0946 \\ 
\multicolumn{1}{c|}{} & Dual-LSTM \cite{DUAL-LSTM} & \multicolumn{1}{c}{0.0885} & \multicolumn{1}{c}{0.0636} & 0.0874 \\
\multicolumn{1}{c|}{} & DeTransformer \cite{1} & \multicolumn{1}{c}{0.0764} & \multicolumn{1}{c}{0.0613} & \textbf{0.0705} \\
\multicolumn{1}{c|}{} & \textbf{De-SaTE} & \multicolumn{1}{c}{\textbf{0.0330}} & \multicolumn{1}{c}{\textbf{0.0080}} & 0.090 \\
\hline
\specialrule{0.1em}{0.05em}{0.05em}
\end{tabular}}
\end{table}

\section{Conclusion and Future Work}
This work uses a denoising framework to filter out noise from the \ac{NASA} and \ac{CALCE} lithium-ion battery data to estimate the \ac{RE}, \ac{MAE}, and \ac{RMSE} metrics. In addition to usual modeling of Gaussian noise, this study extends the recent literature to model multiple types of noise distributions. The findings show that Poisson noise produces a lower \ac{RE} of 0.033 over the other noises for the \ac{CALCE} dataset. However, Gaussian noise yields enhanced performance across \ac{RE} and \ac{MAE} for the \ac{NASA} dataset. The proposed architecture produces lower \ac{RE}, \ac{MAE}, and \ac{RMSE} compared to past work.

In future, the response of the proposed architecture to adversarial attacks may be proposed and defense strategies may be devised accordingly, thereby adding to its robustness. As a next step, a more traditional encoder-decoder model might be introduced to extend the present capabilities and advance the understanding of battery health prognostics.

\section*{Acknowledgment}
The research reported in this publication was supported by the Division of Research and Innovation at San Jos\'e State University under Award Number 23-UGA-08-044. The content is solely the responsibility of the author(s) and does not necessarily represent the official views of San Jos\'e State University.
\bibliographystyle{IEEEtran}
\bibliography{mybib}

\newpage

\appendices

    \section{Related Work}
\label{related-work}
 \ac{Li-ion} batteries are a ubiquitous power source in industries like electric cars, drone technology, and various other applications, offering efficient energy storage. Battery prognosis has seen rising trends with the widespread use of \ac{Li-ion} batteries in the industry. Recent literature \cite{shift} has seen drift from model-based techniques to deep-learning based models. 

\subsection{Model Based Methods}
Model-based methods attempt to set up mathematical or physical models to describe degradation processes, and update model parameters using measured data \cite{11, 12, 13}. However, these methods require accurate knowledge of the battery's internal structure and operating conditions, which can be challenging to obtain in practice.

\vspace{0.25cm}
\subsubsection{Physical Model}
Physical model quantifies the factors that influence a battery’s performance. This approach usually focuses on the specific physical and chemical phenomena occurring during utilization \cite{14}. Physics-based models rely on mathematical equations to describe the battery's physical attributes and controlling principles. 
\vspace{0.25cm}
\subsubsection{Electrochemical Model}
Electrochemical models are based on precise mathematical models of electrochemical processes that occur within the battery, such as chemical reactions, lithium ion and electron movement, and heat impacts  \cite{15}. However, due to the complexity and non-linearity of battery behavior, as well as the challenge of precisely describing the electrochemical processes within the battery, establishing accurate electrochemical-based models can be difficult. 
\vspace{0.25cm}
\subsubsection{Adaptive Filter Method}
Adaptive filter is a digital filter whose coefficient varies with the target for the filter to converge to the optimal state \cite{32, 33}. Multiple research studies have demonstrated that adaptive filters function well in \ac{RUL} estimation of \ac{Li-ion} batteries.

\vspace{0.25cm} 
\vspace{0.25cm}
\subsubsection{Stochastic Process Methods}
Stochastic process methods are based on the notion that battery degradation is a stochastic process that can be modeled using probabilistic methods. The advantage of stochastic approaches is that they can represent the unpredictability and uncertainty inherent in the battery deterioration process. But they may need more complicated modeling and computational techniques\cite{44}. 
\subsection{Data Driven Methods}
Data-driven methods rely extensively on analyzing the battery's operating data to estimate its degradation level and predict when it will reach the end of its useful life. 
Such a method can directly mime the degradation information of lithium-ion battery through historical data, and there is no need to establish a specific mathematical model \cite{16, 17}. 
\subsubsection{Traditional Machine Learning}
This section provides an overview of traditional machine learning methods for \ac{RUL} estimation of \ac{Li-ion} batteries. The techniques that utilize \ac{RVM} \cite{19, 20, 21} achieve efficient online training for updating the model using battery data. However, they pose a memory consumption issue with increased model complexity. Multiple ensemble models \cite{21, 22, 23, 24} have been proposed to improve prediction accuracy. Over the years, traditional machine learning methods have been widely used for \ac{RUL} estimation of \ac{Li-ion} batteries. While these methods have shown promising results, they also have their limitations. Overall, these methods offer a range of approaches to address the challenge of \ac{RUL} estimation and can be useful in various applications, but careful consideration is needed when selecting the most appropriate method for a particular scenario.


\subsubsection{The Advent of Deep Learning}

Deep learning models automatically learn relevant features from raw data, capture complex relationships between input features and output targets, and generalize well to new data. \ac{LSTM} models and their variants \cite{25, 26, 27, 31} can extract multi-dimensional features and estimate \ac{RUL} with high precision. Zhang et al. \cite{28} proposed an online estimation method that combines partial incremental capacity with an \ac{ANN} for estimating battery \ac{SOH} and \ac{RUL}. \ac{TCN} \cite{29} that uses causal and dilated convolution techniques to capture local capacity regeneration has improved prediction accuracy. Overall, the shift from traditional machine learning to deep learning has led to significant improvements in the accuracy and robustness of \ac{RUL} estimation methods for \ac{Li-ion} batteries. However, deep learning models can be computationally expensive to train and require large amounts of data, which may limit their applicability in some domains.

In recent years, the landscape of \ac{RUL} estimation for \ac{Li-ion} batteries has undergone a remarkable transformation with the emergence of transformer-based models \cite{1}. This paradigm shift can be attributed to the extraordinary capabilities of transformers in processing sequential data efficiently, rendering them exceptionally well-suited for intricate time-series forecasting tasks such as \ac{RUL} estimation. Transformers have risen to prominence due to their innate prowess in capturing long-term dependencies within data, endowing them with the capacity to model the intricate and evolving degradation patterns of Li-ion batteries with precision.

  \section{Evaluation Metrics}
  \label{SecondAppendix}
    \subsection{Relative Error (\ac{RE}):}
    Relative Error measures the relative difference between predicted and actual values and is represented as follows:
    
    \begin{equation}
        RE = \frac{|Y - \hat{Y}|}{|Y|}
    \end{equation}

    where,
    \(\hat{Y}\) represents the predicted value, and
    \(Y\) represents the actual value.

    \subsection {Root Mean Square Error (\ac{RMSE}):}
    \ac{RMSE} calculates the square root of the mean of the squared differences between predicted and actual values:
    
    \begin{equation}
    RMSE = \sqrt{\frac{1}{n}\sum_{i=1}^{n}(\hat{Y}_i - Y_i)^2}
    \end{equation}
    
    where,
    \(\hat{Y}_i\) represents the predicted value for the \(i\)-th sample,
    \(Y_i\) represents the actual value for the \(i\)-th sample, and
    \(n\) is the total number of samples.
    
    \subsection{Mean Absolute Error (\ac{MAE}):}
    \ac{MAE} calculates the mean of the absolute differences between predicted and actual values:
    
    \begin{equation}
    MAE = \frac{1}{n}\sum_{i=1}^{n}| \hat{Y}_i - Y_i |
    \end{equation}
    
    where,
    \(\hat{Y}_i\) represents the predicted value for the \(i\)-th sample,
    \(Y_i\) represents the actual value for the \(i\)-th sample, and \(n\) is the total number of samples.

    \section{High-level overview of the experimental framework}
\begin{figure}[htbp]
      \centering  \includegraphics[width=1\columnwidth]{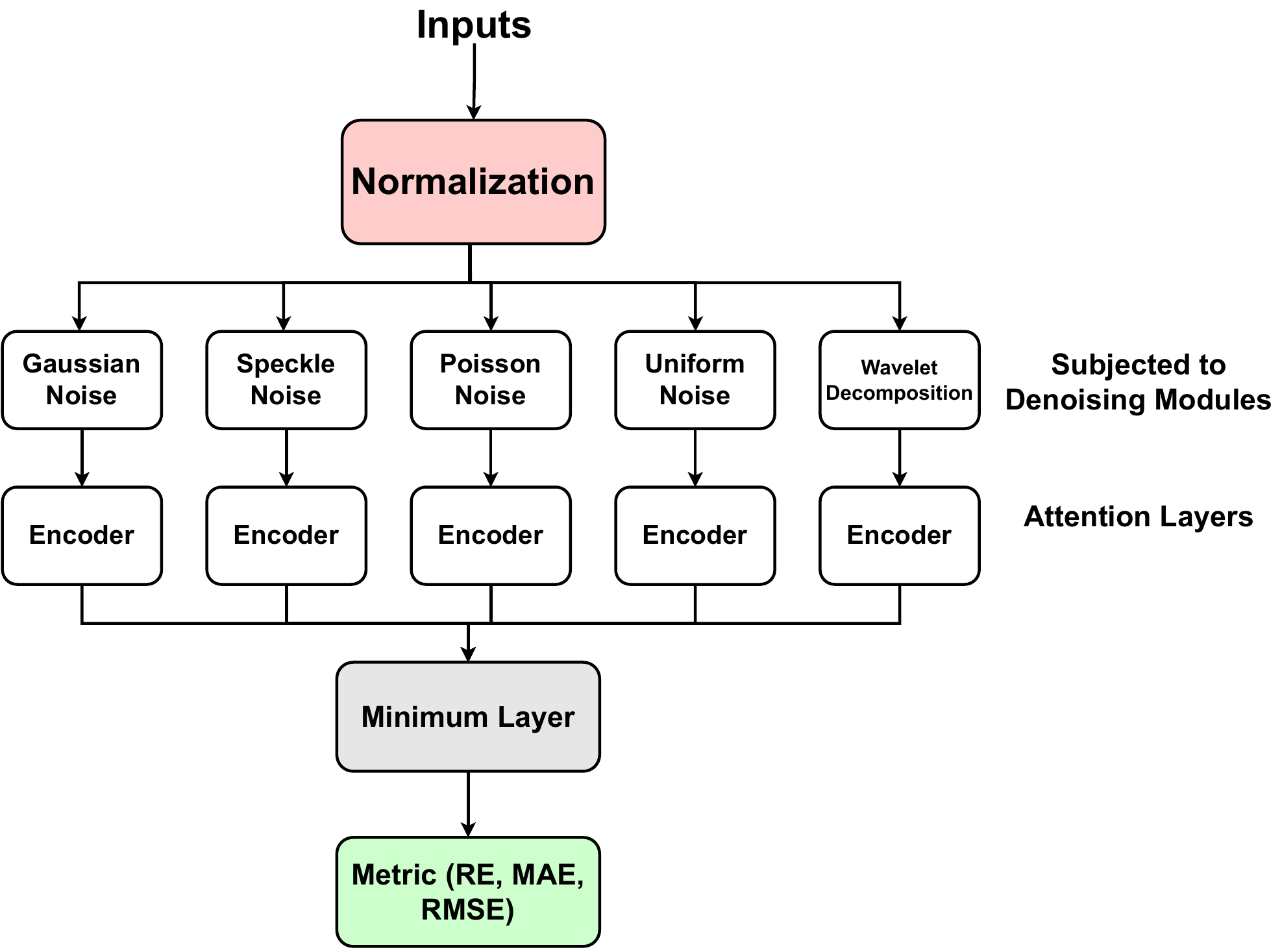}
        \caption{High-level overview of the experimental framework}
        \label{high_level_ovr}
    \end{figure}

    The initial phase of the experimental framework is shown in Fig. \ref{high_level_ovr} in which the input data is standardized to establish a uniform scale, ensuring coherence. Following that, the denoising modules are used, which are designed to refine the data by filtering out extraneous noise, thereby improving the dataset's integrity. The transformer-encoder component extracts intricate features from refined data and computes critical metrics like RE, MAE, and RMSE. This sequential protocol, encapsulates the series of operations that form the foundation of our experimental pipeline.
    
    \section{Abbreviations}
  \noindent Below we list all the abbreviations used in this paper.
  \label{FirstAppendix}
  \begin{acronym}[AAAAA]
       \acro{Li-ion}{Lithium-ion}
       \acro{RUL}{Remaining Useful Life}
       \acro{PHM}{Prognostics and Health Management}
       \acro{EOL}{End of Life}
       \acro{RVM}{Relevance Vector Machine}
       \acro{SOH}{State of Health}
       \acro{NASA}{National Aeronautics and Space Administration}
       \acro{CALCE}{Center for Advanced Life Cycle Engineering}
       \acro{LSTM}{Long Short Term Memory}
       \acro{ANN}{Artificial Neural Network}
       \acro{TCN}{Temporal Convolution Network}
       \acro{CNN}{Convolutional Neural Network}
       \acro{RNN}{Recurrent Neural network}
       \acro{PDF}{Probability Density Function}
       \acro{RE}{Relative Error}
       \acro{RMSE}{Root Mean Squared Error}
       \acro{MAE}{Mean Absolute Error}
       \acro{DWT}{Discrete Wavelet Transform}
       \acro{IDWT}{Inverse Discrete Wavelet Transform}
       \acro{NoL}{Number of layers}
       \acro{HD}{Hidden Dimensions}
       \acro{LR}{Learning rate}
    \end{acronym}

\end{document}